\documentclass{article}
\usepackage{ijcai22}

\usepackage{times}
\usepackage{soul}
\usepackage{url}
\usepackage[hidelinks]{hyperref}
\usepackage[utf8]{inputenc}
\usepackage[small]{caption}
\usepackage{graphicx}
\usepackage{amsmath}
\usepackage{amsthm}
\usepackage{booktabs}
\usepackage{algorithm}
\usepackage{algorithmic}
\usepackage{epsfig}
\usepackage{amssymb}
\usepackage{multirow}
\usepackage{array}
\usepackage{color}
\usepackage{makecell}
\usepackage{enumitem}
\usepackage{float}
\usepackage{setspace}
\usepackage{cases}
\usepackage[explicit]{titlesec}
 
\titlespacing*{\section}{0pt}{0.5ex plus .0ex minus .0ex}{.2ex plus .0ex}
\titlespacing*{\subsection}{0pt}{0.5ex plus .0ex minus .0ex}{.2ex plus .0ex}
\titlespacing*{\subsubsection}{0pt}{0.2ex plus .0ex minus .0ex}{.2ex plus .0ex}

\title{Plane Geometry Diagram Parsing}

\author{
Ming-Liang Zhang$^{1,2,}$\footnote{Contact Author}\and
Fei Yin$^{1,2}$\and
Yi-Han Hao$^{1,3}$\And
Cheng-Lin Liu$^{1,2}$\\
\affiliations
$^1$National Laboratory of Pattern Recognition, Institute of Automation of Chinese Academy of Sciences\\
$^2$School of Artificial Intelligence, University of Chinese Academy of Sciences\\
$^3$School of Electronic Information Engineering, Beijing Jiaotong University\\
\emails
zhangmingliang2018@ia.ac.cn,
fyin@nlpr.ia.ac.cn,
20120004@bjtu.edu.cn, 
liucl@nlpr.ia.ac.cn
}

\begin{document}

\maketitle

\begin{abstract}
Geometry diagram parsing plays a key role in geometry problem solving, wherein the primitive extraction and relation parsing remain challenging due to the complex layout and between-primitive relationship. In this paper, we propose a powerful diagram parser based on deep learning and graph reasoning. Specifically, a modified instance segmentation method is proposed to extract geometric primitives, and the graph neural network (GNN) is leveraged to realize relation parsing and primitive classification incorporating geometric features and prior knowledge. All the modules are integrated into an end-to-end model called PGDPNet to perform all the sub-tasks simultaneously. In addition, we build a new large-scale geometry diagram dataset named PGDP5K with primitive level annotations. Experiments on PGDP5K and an existing dataset IMP-Geometry3K show that our model outperforms state-of-the-art methods in four sub-tasks remarkably. Our code, dataset and appendix material are available at \url{https://github.com/mingliangzhang2018/PGDP}. 
\end{abstract}

\section{Introduction}

Automatic geometry problem solving is a long-standing problem and has important applications in the intelligent education field \cite{Chou1996,Seo2015,Amini2019}. The problem involves text parsing, corresponding diagram parsing and logical reasoning. Previous research works \cite{Sachan2017,Sachan2020} mainly concentrated on text parsing and logical reasoning, but little attention has been paid to diagram parsing \cite{Seo2014,Lu2021}. Geometry diagrams carry rich information about the geometry problem, which can provide crucial cues to aid problem solving. In this work, we focus on plane geometry diagram parsing (PGDP) and propose a powerful diagram parser.

\begin{figure}[t]
    \begin{center}
    \includegraphics[width=1.0\columnwidth]{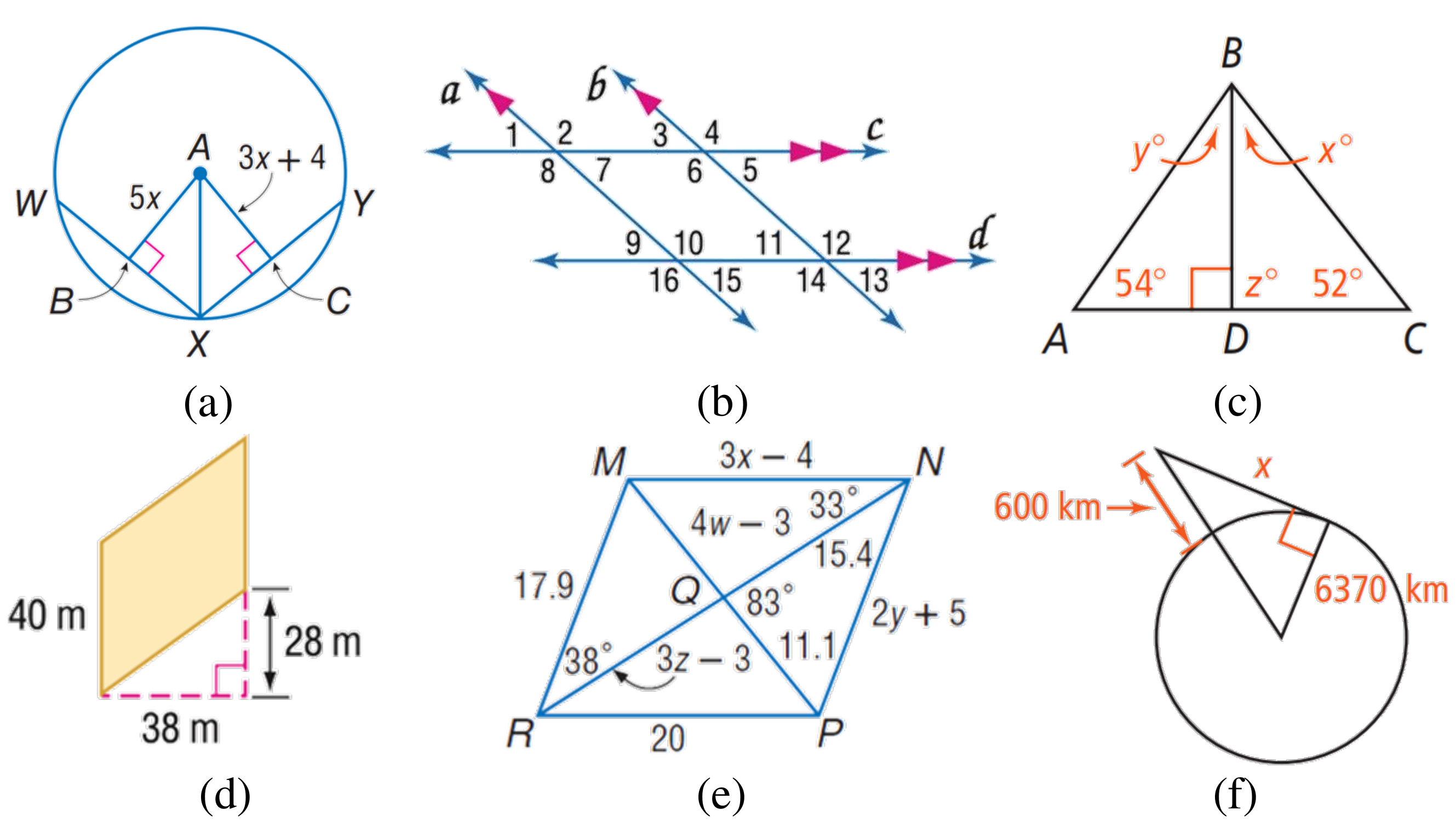} 
    \end{center}
    \caption{Examples of plane geometry diagram.}
    \label{cover}
\end{figure}

Generally, the PGDP task involves identifying and locating visual primitives in the diagram and discovering relationships among them. As shown in Figure \ref{cover}, a geometry diagram consists of various types and layouts of geometry, symbols and texts, and these visual primitives are semantically related to each other in various ways. Due to the diversity of style and the interference of primitives, traditional methods, such as Hough transform and Freeman chain-code \cite{Pratt2001}, perform poorly in geometric primitive extraction. Meanwhile, the spatial, structural and semantic relations among primitives cannot be parsed correctly by simple rule-based methods \cite{Seo2014,Lu2021}. Therefore, great efforts are needed for geometric primitive extraction and between-primitive relationship parsing.

We cast geometric primitive extraction as an instance
segmentation problem. Geometric primitives such as lines and arcs are often slender and overlapped. Thus, bounding box based instance segmentation methods \cite{He2018,Neven2019,Ying2021} are not suitable for this task. Our proposed PGDP framework instead employs a geometric segmentation module (GSM), consisting of a semantic segmentation branch and a segmentation embedding branch, to cluster multi-class primitive instances at pixel level, so as to overcome the issues stated above.

\begin{figure*}[htbp]
    \begin{center}
    \includegraphics[width=1.85\columnwidth]{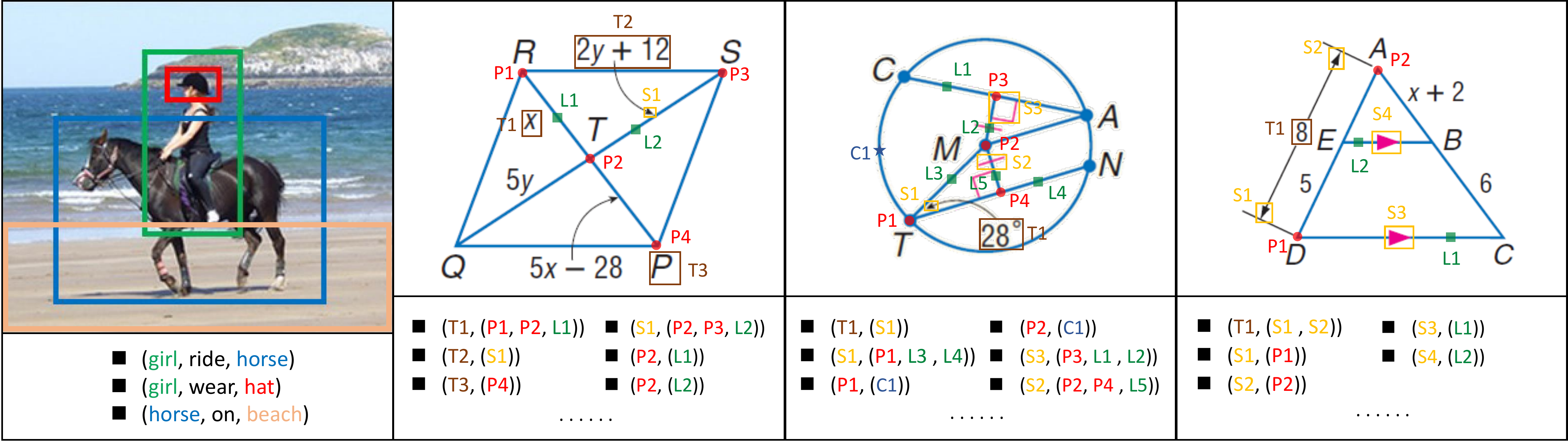} 
    \end{center}
    \caption{Comparison between tasks of SGG (first image) and PGDP (next three images). Relation tuples are shown below each image. `P\#',`L\#',`C\#',`T\#' and `S\#' denote instances of point, line, circle, text and symbol, respectively.}
    \label{compare with SGG}
\end{figure*}

For primitive relation parsing, we model PGDP as a special scene graph generation (SGG) problem \cite{Xu2017,Liu2021}. In contrast to ordinary SGG, as shown in Figure \ref{compare with SGG}, PGDP deals with graphs with heterogeneous nodes and multi-edges associated, when primitives and their relations are seen as nodes and edges, respectively. To optimize reasoning incorporating geometry prior knowledge, we adopt a GNN module (GM) aggregated with visual, spatial and structural information to predict the primitive relation and identify the text class simultaneously. 

Integrating with GSM and GM, we present the deep learning model for PGDP called PGDPNet. The PGDPNet is trained end-to-end so as to optimize the overall primitive extraction and relation reasoning performance. Also, to facilitate the research of PGDP, we build a new large-scale geometry diagram dataset named PGDP5K, labeled with annotations of primitive locations, classes and their relations. Experiments on PGDP5K and an existing dataset IMP-Geometry3K demonstrate that our method can boost the performance of primitive detection, relation parsing and geometry formal language generation prominently, compared to state-of-the-art methods, and consequently improves the accuracy of geometry problem solving.

The contributions of this work are summarized in three folds: (1) We propose the PGDPNet, the first end-to-end deep learning model for explicit geometry diagram parsing. (2) We build a large-scale dataset PGDP5K, containing fine-grained annotations of primitives and relations. (3) Our method demonstrates superior performance of geometry diagram parsing, outperforming previous methods significantly.

\section{Related Work}
Automatic analysis of geometry diagrams has been studied in two main aspects: primitive extraction and relation reasoning. As to primitive extraction, traditional methods such as Hough transform and its improved methods \cite{Pratt2001} are still adopted in most recent geometry diagram parsing works \cite{Seo2014,Seo2015,Gan2018,Lu2021} for their simplicity and efficiency. However, in the scenes of complex layout and multi-primitive interference, traditional methods inevitably suffer severe performance degradation. Deep learning based geometric primitive extraction methods \cite{Huang2018,Zhou2019} have been proposed recently. Nevertheless, they only focus on one type of geometric primitive such as straight line in nature scenes. Research works about geometric relation reasoning of diagrams are undergoing. Some methods \cite{Seo2014,Lu2021} use greedy or optimization strategies based on distance and content rules, but cannot parse complicated between-primitive relations correctly. Our work, inspired by the SGG task \cite{Xu2017,Liu2021,Guo2021}, reasons primitive relations with the GNN model \cite{Romero2018,Ye2020}. The detailed comparison between these two tasks will be described in Section 3.2. To sum up, our work proposes a more powerful geometry diagram parser with a novel and effective scheme for diagram primitive extraction and reasoning.


\section{Preliminary}
    Before the description of problem formulation and solution, we introduce the terms involved in PGDP. The basic element in plane geometry diagram is called \textit{primitive}, which is generally categorized into \textit{geometric primitive} and \textit{non-geometric primitive}. The main categories of geometric primitives are \textit{point}, \textit{line} and \textit{circle} (\textit{arc}), while non-geometric primitives include \textit{text} and \textit{symbol}. \textit{Predicate} is the general term of geometric shape entity, geometric relation or arithmetic function. \textit{Proposition} is the logical expression combined with predicate and primitive. A set of propositions makes up the \textit{geometry formal language} of a diagram.
\subsection{Task Formulation}
The PGDP consists of three fundamental sub-tasks: (1) detection and identification of primitives; (2) building basic relationships among primitives; (3) generating the geometry formal language. In this work, we model PGDP as special SGG, which is formulated in the following form:
\begin{equation}
    \begin{aligned}
    O &=\left\{O_{geo}, O_{sym}, O_{text}\right\} , \\
    B &=\left\{B_{geo}(\text{mask}), B_{sym}(\text{box}), B_{text}(\text{box})\right\} , \\
    R &=\left\{R_{geo2geo}, R_{text2geo}, R_{sym2geo}, R_{text2sym}\right\} ,
    \end{aligned}
\end{equation}
where $O$ is the primitive set, the subscripts geo, sym, text stand for geometric primitive, symbol and text, respectively; $B$ is the primitive position set (geometric and non-geometric primitives are represented by mask and bounding box); $R$ is the primitive relation set among geometric primitive, symbol and text. $G\!\!=\!\!\{O,\!B,\!R\}$ constitutes a special scene graph. The parsing on image $I$ is aimed to predict the constituents of $G$:
\begin{equation}
    \begin{aligned}
        \mathcal{P}\left(G\mid I,K\right)&= \mathcal{P}\left(B\mid I \right) \cdot
                \mathcal{P}\left(\left\{O_{geo},O_{sym}\right\}\mid I,B\right) \\
        & \cdot \mathcal{P}\left(R,O_{text} \mid\left\{O_{geo},O_{sym}\right\},B,I,K\right)
    \end{aligned} ,
\end{equation}
where $K$ denotes the knowledge graph of geometry; $\mathcal{P}\left(B\mid I \right) \cdot \mathcal{P}\left(\left\{O_{geo},O_{sym}\right\}\mid I,B\right)$ refers to the steps of object detection and instance segmentation for obtaining the primitive positions and classes of symbol and geometric primitive; $\mathcal{P}\left(R,O_{text} \mid\left\{O_{geo},O_{sym}\right\},B,I,K\right)$ stands for relation inference to acquire the primitive relationship and text class. At last, we generate the proposition set to form the readable description language.  
\begin{figure}[t]
    \begin{center}
    \includegraphics[width=0.8\columnwidth,trim=0 5 0 15,clip]{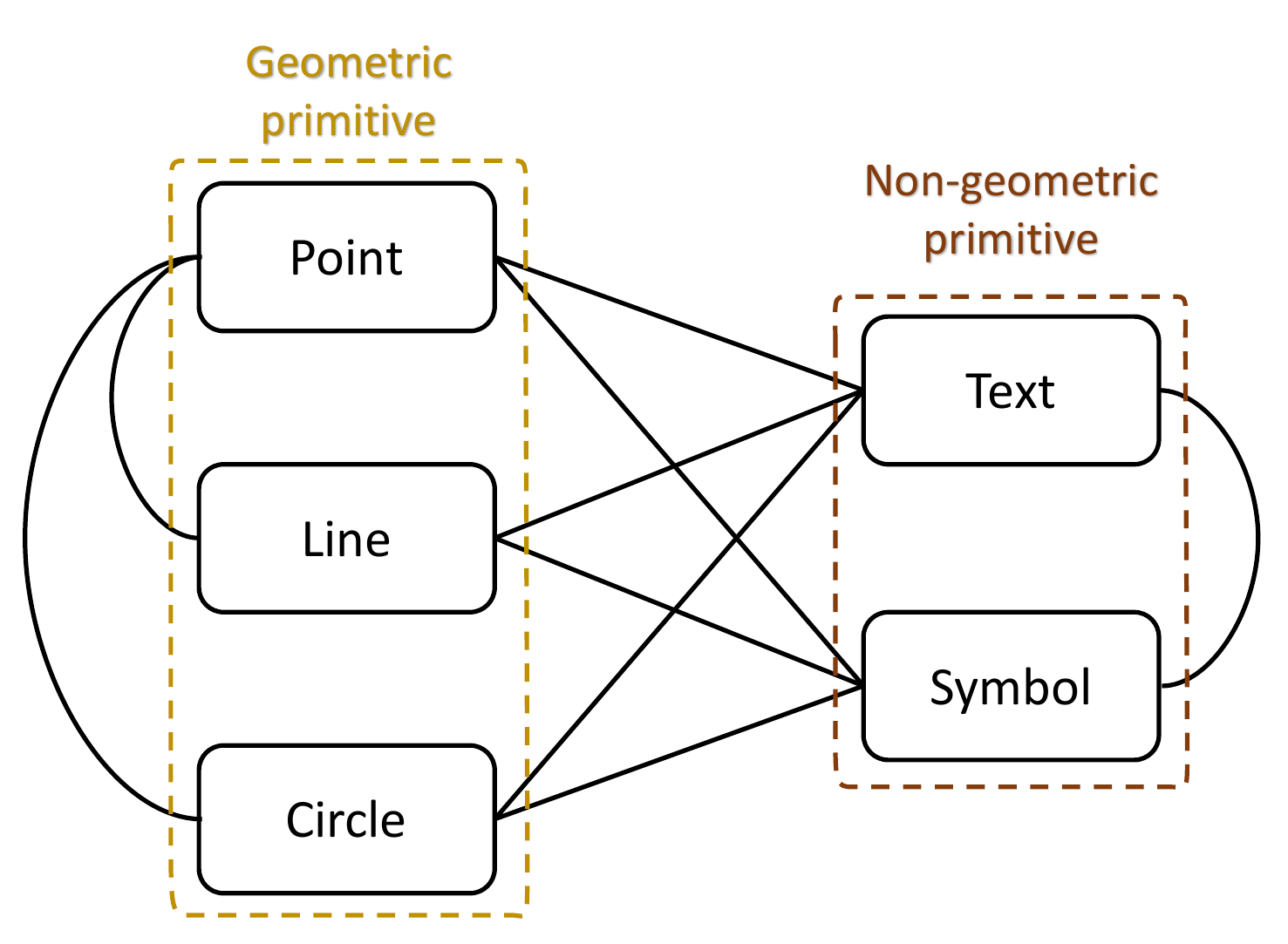} 
    \end{center}
    \caption{Primitive relationship graph of plane geometry diagram.}
    \label{graph}
\end{figure}

\begin{figure*}[htbp]
    \begin{center}
    \includegraphics[width=2.0\columnwidth]{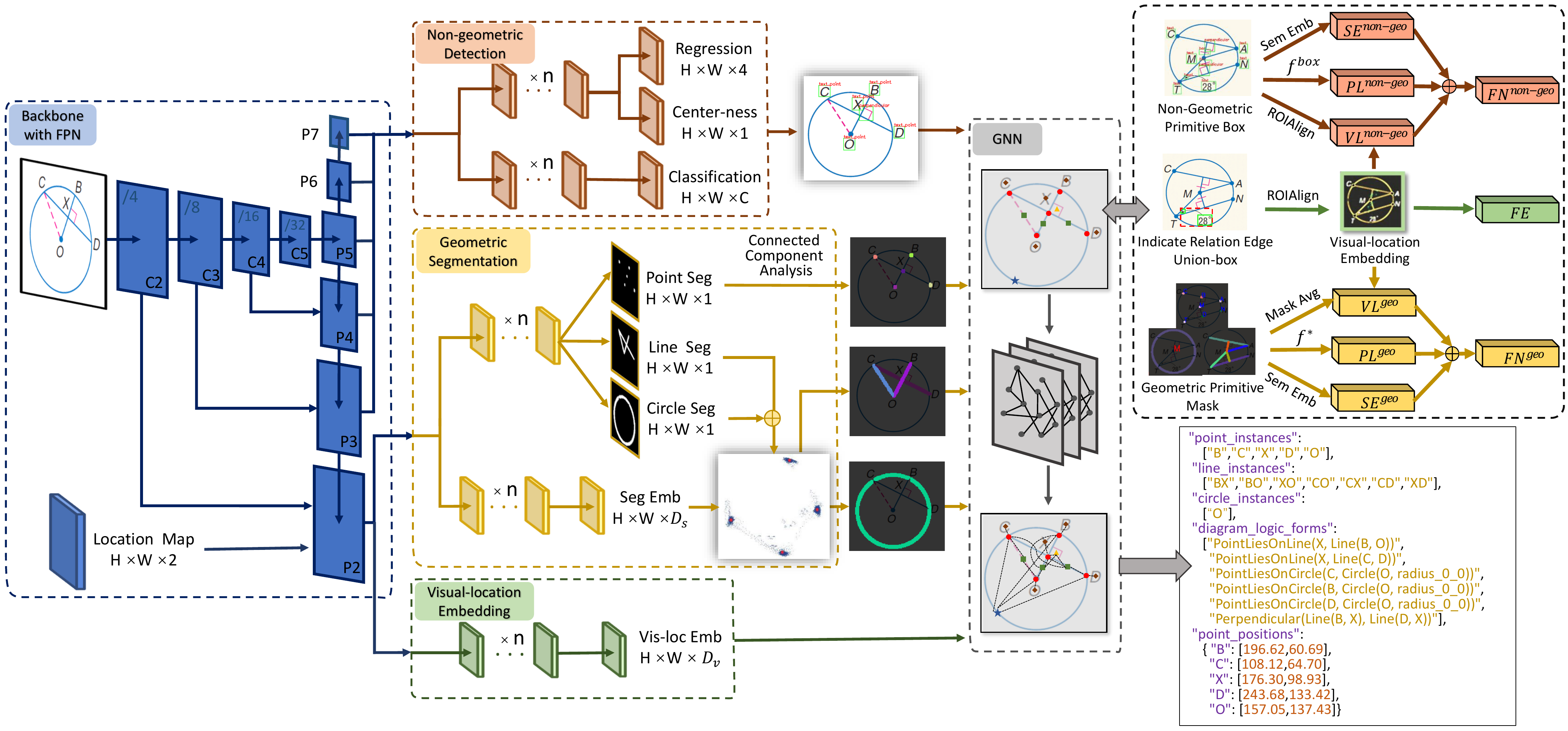} 
    \end{center}
    \vspace{-0.2cm}
    \caption{Overview of our proposed PGDPNet.}
    \label{network}
\end{figure*} 
\subsection{Comparison with SGG}
PGDP is different from general SGG in two respects as shown in Figure \ref{compare with SGG}. First, SGG obtains coarse box positions of targets from nature scene images through object detection, while PGDP is aimed to get the fine-grained masks of geometric primitives through instance segmentation, because they largely overlap with each other in box. Second, the SGG constructs the relationship graph in the form of subject-predicate-object triplets, which has no necessary dependency between the predicate and subject/object classes, while the between-object relationship in PGDP is mostly geometric, and specific relations (predicate) could be inferred according to primitive classes (subject/object) and prior knowledge, do not require an extra classification step.

\section{PGDP5K Dataset}
Although several datasets \cite{Seo2015,Sachan2017,Lu2021,Chen2021} for solving geometry problems have been proposed, there is no dataset focusing on PGDP. To facilitate research in geometry problem solving, we build a new large-scale and fine-annotated plane geometry diagram dataset named PGDP5K\footnote{ \url{http://www.nlpr.ia.ac.cn/databases/CASIA-PGDP5K}}.
  
\subsection{Statistics} 
The PGDP5K dataset contains 5,000 diagram samples, consisting of 1,813 non-duplicated images from the Geometry3K dataset and other 3,187 images collected from three popular textbooks across grades 6-12 on mathematics curriculum websites\footnote{\url{https://www.mheducation.com/}}. We randomly split the dataset into three subsets: train set (3,500), validation set (500) and test set (1,000). In contrast to previous datasets, diagrams in PGDP5K have more complex layouts such as multiple classes of primitives and complicated primitive relations, which make our dataset more challenging. Specifically, we divide the geometric primitive, text and symbol into 3, 6 and 16 classes respectively, and Appendix A displays several instances of each class primitive. Some classes of primitives have great within-class style variations. The Appendix B shows distributions of shape, symbol, text and relation. 

\subsection{Annotations} 
The annotations of PGDP5K dataset include three types: geometric primitive, non-geometric primitive and primitive relation. These annotations can generate geometry formal language automatically and uniquely. As to geometric primitives, we annotate their pixel positions and uniform pixel widths. For non-geometric primitives, bounding box, symbol class, text class and text content are labeled. As to primitive relations, we construct a relation graph of elementary relationships among primitives exhibited in Figure \ref{graph}, where we only construct relations between point and line, point and circle for relations of geometric primitives, because other high-level relations among geometric primitives can be derived from these two basic relations. A two-tuple with multiple entities is used to represent one relationship as demonstrated in Figure \ref{compare with SGG}. Compared with the triplet of SGG, we take point, symbol and text as subjects, and serve other related primitives as objects, neglecting the relation class term. Finally, we define geometry proposition templates of basic relations listed in Appendix C. For more annotation details, please refer to the website of PGDP5K dataset. In addition, we re-annotate diagrams of the Geometry3K in our way and rename it IMP-Geometry3K.

\section{Model}
The proposed PGDPNet depicted in Figure \ref{network} is presented hereon in detail, focusing on the geometric primitive segmentation and the  primitive relation parsing.

\subsection{Backbone Module (BM)}
A typical FPN architecture \cite{Lin2017} is used as the BM for visual feature extraction. The FPN layers P3-P7 are exploited for text and symbol detection, and the FPN layer P2 embedded with the location maps is shared by the geometric segmentation module (GSM) and the visual-location embedding module (VLEM). Visual features mixed with spatial information will facilitate model learning of follow-up tasks. 

\subsection{Non-geometric Detection Module (NDM)} 
The non-geometric primitives, symbol and text, are detected in NDM. Given the diverse size scales of text and symbol in geometry diagrams, an anchor-free detection method FCOS \cite{Tian2020} is utilized to avoid the setting of prior anchors and improve the detection speed. This module consists of regression, center-ness and classification branches. The loss in training is  $L_{F\!C\!O\!S}=L_{reg}+L_{cns}+L_{cls}$, where $L_{reg}$, $L_{cns}$ and $L_{cls}$  are the losses of three branches.

\subsection{Geometric Segmentation Module (GSM)}

Due to complex layouts and elongated shapes, traditional methods and current instance segmentation approaches based on bounding boxes are all not suitable for geometric primitive extraction. We propose a new instance segmentation method, where two branches, semantic segmentation branch and segmentation embedding branch, together implement the multi-class instance segmentation of geometric primitives. The semantic segmentation branch performs binary segmentation with the weighted binary cross-entropy (BCE) loss:
\begin{equation}
L_{bs*}=\frac{-w_{*}}{M_{map}} \sum_{i=1}^{M_{map}}y_{i}^{*}\!\log\left(p_{i}^{*}\right)\!+\!\left(1\!-\!y_{i}^{*}\right)\log\left(1\!-\!p_{i}^{*}\right) ,
\end{equation} 
where * denotes the primitive class, $w$ is the weight ratio for balancing positive and negative class pixels, empirically as $w_{p}\!=\!5, w_{l}\!=\!1, w_{c}\!=\!4$, and $M_{map}$ is the pixel number of segmentation map.
Then the segmentation loss of all classes is $L_{bs}\!=\!L_{bsp}\!+\!L_{bsl}\!+\!L_{bsc}$.
The discriminative loss \cite{DeBrabandere2017,Neven2018} is used in the segmentation embedding branch to better differentiate instances:
\begin{align}
\left\{    
	\begin{aligned}
    L_{dist}\!&=\!\frac{1}{N_{lc}\left(N_{lc}\!-\!1\right)}\!\!\sum_{n_{1}=1}^{N_{lc}}\!\!\sum_{n_{2}=1 \atop {n_{2}\neq n_{1}}}^{N_{lc}}\!\!\left[2\delta_{d}\!-\!\left\|\mu_{n_{1}}\!\!-\!\mu_{n_{2}}\right\|\right]_{+}^{2},\\[-3mm]
    L_{var}\!&=\!\frac{1}{N_{lc}}\sum_{n=1}^{N_{lc}} \frac{1}{M_{n}} \sum_{i=1}^{M_{n}}\left[\left\|\mu_{n}\!-\!x_{i}\right\|\!-\!\delta_{v}\right]_{+}^{2},
	\end{aligned}
\right.
\end{align}
where $N_{lc}=N_l+N_c$ is the sum of instance number of line and circle, $M$ is the pixel instance number, $x$ is the pixel embedding, $\mu$ is the center of instance embedding. In default, the threshold $\delta_{d}$ of center distance is set as 1.5 and threshold $\delta_{v}$ of embedding radius is set as 0.5. The line instances and circle instances are learned simultaneously to obtain more distinctive features. In contrast, due to inherent instance separation in space, point instances are acquired just by connected component analysis according to the results of semantic segmentation, reducing difficulty of model learning. Eventually, the whole loss of GSM is $L_{ins}\!=\!L_{bs}\!+\!L_{dist}\!+\!L_{var}$. 

\subsection{GNN Module (GM)}


\begin{table*}[t]
    \small
    \renewcommand\arraystretch{1.1}
    \centering
    \begin{tabular}{l|ccc}
    \toprule
      & InterGPS & PGDPNet w/o GNN & PGDPNet\\
    \midrule
      Non-geometric Primitive Detection & Object Detection & Object Detection & Object Detection \\
      Geometric Primitive Detection & Hough Transform & Instance Segmentation & Instance Segmentation \\
      Text Classification & Content Rules & Detection Classification & Joint Classification \\
      Relation parsing	& Distance Rules & Distance Rules & GNN \\
      Has Parsing Position & Yes & No (localizable) & No (localizable) \\
      Is End-to-End	& No & No &	Yes \\
    \bottomrule
    \end{tabular}
    \caption{Functions of comparison methods.}
    \label{benchmarks}
\end{table*}

After obtaining the primitives, the relationship among primitives is reasoned by the GM. Before that, the VLEM unifies all primitive features and works as the initialization of GM. We treat primitives as nodes and primitive relations as edges to compose a primitive relation graph, represented as $G\!\!=\!\!\{V\!\!=\!\!\{V^{geo}, V^{N\!on\!-\!geo}\},E\!\!=\!\!\{e_{ij}|K\}\}$, where $V^{geo}$, $V^{N\!on\!-\!geo}$ and $E$ denote the node sets of geometric and non-geometric primitive, and the edge set. The cardinality of geometric node set is $|V^{geo}|\!=\!N_{p}\!+\!N_{l}\!+\!N_{c}$, where $N_p$ is the instance number of point. The original relation graph is a difficult hyper-graph with heterogeneous nodes (mask and box) and multi-edge relationship. For efficient solution, we transform the graph to a simple isomorphic graph and then construct the sparse graph like the one in Figure \ref{graph}. Specifically, we only connect nodes that may have relations according to geometric prior knowledge $K$, and then categorize edges to determine final results. 

The initial features of nodes are represented by the fusion of visual-location embedding $V\!L$, parsing position feature $P\!L$ and class semantic feature $S\!E$. For visual-location embedding, we transform mask features of variable shapes into vector features of fixed length by the mask average:
\begin{equation}
    V\!L^{geo}_{i}=B^{T}\dfrac{\text{mask}_{i}}{\|\text{mask}_i\|_1} ,
\end{equation}
where $B$ is the visual-location embedding map. The mask average can also reduce the influence of error caused by inaccurate segmentation. In addition, we employ the RoIAlign method \cite{He2018} to normalize multi-scale box features as vector features of the same length:
\begin{equation}
    V\!L_{i}^{non-geo}=\text{RoIAlign}(B, \text{box}_{i}) .
\end{equation}
Besides, parsing position in plane space is a significant feature of geometric primitives. We incorporate it into the GM to further facilitate relation building, formulated as:
\begin{equation}
    P\!L_{i} = f^{*}(pr_{i}^{*}),\; pr^{*}\!=\!\left\{\begin{aligned}
    &[x,y], &*=\text{point} ,\\[-1.2mm]
    &[x_1, y_1, x_2, y_2], &*=\text{line} ,\\[-1.2mm]
    &[x, y, r], &*=\text{circle} ,\\[-1.2mm]
    &[x_1, y_1, x_2, y_2], &*=\text{box} ,
    \end{aligned}
    \right.
    \label{eqVL}
\end{equation}
where $pr^{*}$ is the parsing representation of primitives, while point, line, circle and box denotes as a point, two endpoints, center with radius, and top-left point with bottom-right point, respectively, $f^{*}$ is a network module of  two fully-connected layers with ReLU activation. In that way, the final node feature is formulated as $F\!N_{i}^{\&}\!=\!V\!L_{i}^{\&}\!+\!P\!L_{i}^{\&}\!+\!S\!E_{i}^{\&}, i\!\!=\!\!1\!\cdots\!|V^{\&}|$, where $\&$ is the primitive class $geo$ or $non\!\!-\!\!geo$. 

The edge features are only generated for arrow indication relations, and the rest are aggregated through layer propagation of GNN. Because of elongated shape, the detection performance of arrows is less satisfactory. Instead of detecting boundary boxes of arrows, we use the union box of head box and corresponding text box to represent the relation:
\begin{equation}
   F\!E_{ij}\!\!=\!\!\left\{\!\begin{aligned}
    \text{RoIAlign}(B, &\text{box}_{i}\!\cup\!\text{box}_{j}), (v_i,\!v_j\!)\!\!=\!\!(\text{head},\!\text{text}\!) , \\
    &\textbf{0}, \qquad\qquad \quad \text{others} .
    \end{aligned}
    \right.
\end{equation}
Although union box cannot enclose the whole arrow in some cases, it works well in experiments because the feature box has a larger receptive field than its own size with the FPN. 

The GM performs two sub-tasks. The first is predicting the edge class to judge whether existing relationship between nodes, with the loss function as:
\begin{equation}
    L_{edge}=\frac{-1}{|E|} \sum_{i=1}^{|E|}y_{i}\!\log\left(p_{i}\right)\!+\!\left(1\!-\!y_{i}\right)\log\left(1\!-\!p_{i}\right) .
\end{equation}
The second is the fine-grained text classification with CE loss:
\begin{equation}
L_{node}=\frac{-1}{N_t}\sum_{i=1}^{N_t}\sum_{c=1}^{C_{text}}y_{ic}log(p_{ic}) ,
\end{equation}
where $N_t$ is the instance number of text and $C_{text}$ is the text class number. Compared with visual features alone, the combined features including spatial structure information promote fine-grained classification, considering that some texts of different classes are visually identical. As to the architecture of GNN, the edge graph attention network (EGAT) \cite{Guo2021,Ye2020} is employed as the backbone of GM for its excellent reasoning ability among nodes and edges, and the whole loss of GM is $L_{G\!N\!N}\!=\!L_{edge}\!+\!L_{node}$.

\subsection{Training and Testing}
During the training, the model PGDPNet is trained end-to-end with the aggregated loss:
\begin{equation}
    L_{all} = L_{F\!C\!O\!S}+\alpha \cdot L_{ins} + \beta \cdot L_{G\!N\!N} .
\end{equation}
Empirically we set the weight coefficients $\alpha\!=\!\beta\!=\!4$. During testing, according to binary masks obtained from the semantic segmentation branch of GSM, the segmentation embedding branch clusters embedding features to get instances of line and circle by the MeanShift cluster method. The parsing position of instance masks could be located accurately by simple fitting methods due to precise segmentation. The extracted geometric and non-geometric primitives go through the VLEM to generate initial features of GM, then the GM gets relations among primitives via node and edge classification. In the end, geometric propositions are produced according to geometry prior knowledge and language grammar.

\section{Experiments}
\subsection{Experimental Setup}
\subsubsection{Implementation Details} 
We implemented our method using the PyTorch and FCOS framework \cite{Tian2020}. The backbone adopts MobileNetV2 \cite{Sandler2018}. The NDM, GSM and VLEM all use 3 groups of 128-channel convolution layers with corresponding BatchNorm layers. The segmentation embedding dimensionality is 8 and the visual-location embedding dimensionality is 64. The layer number of GM is 5 and the feature dimensionalities of nodes and edges are all set to 64. To improve the diversity of samples, two enhancement strategies, random scale scaling and random flipping, are exploited during the training. We choose the Adam optimizer with an initial learning rate $5e^{-4}$, weight decay $1e^{-4}$, step decline schedule decaying with a rate of 0.2 at 20K, 30K and 35K iterations. We train our model in 40K iterations with batch size of 12 on 4 TITAN-Xp GPUs.
    
\subsubsection{Comparison Methods} 
To evaluate the effects of different modules, we compare three methods: InterGPS, PGDPNet without GNN, and PGDPNet. As described in Table \ref{benchmarks}, these approaches respectively adopt different technologies on sub-tasks, where our PGDPNet is a concise and efficient framework that could be learned end-to-end from datasets.
    
 \subsubsection{Evaluation Protocols} 
 We evaluate the methods at four levels: primitive detection, relation parsing, geometry formal language generation, and problem solving. Considering that some labels of bounding box are loose especially for single-word texts and arrowheads, we set threshold IOU=0.5 to evaluate the non-geometric primitive detection. As to geometric primitive extraction, there are two evaluation manners: one  ({\bf manner 1}) is parsing position evaluation that applies to the Hough transform route and the other ({\bf manner 2}) is mask evaluation designed for the instance segmentation route. We set the distance threshold as 15 consistent with the InterGPS \cite{Lu2021} for the first manner and set IOU as 0.75 by default for the second manner. As to relation parsing, we divide one multivariate relation into multiple binary relations, and evaluate the precision, recall and F1 of binary relation terms. The geometry formal language is characterized by diversity and equivalence, for example, "Angle(P,R,Q)" is equivalent to "Angle(P,R,N)" in Figure \ref{cover}(e). For rationality and fairness of evaluation, we improve the existing evaluation method \cite{Lu2021} focusing on propositions with line and angle. Experimental results are evaluated on four indicators: Likely Same (F1$\geq$50\%), Almost Same (F1$\geq$75\%), Perfect Recall (recall=100\%) and Totally Same (F1=100\%).

\subsection{Primitive Detection}
    To evaluate the effects of NDM and GSM, we give the performance of primitive extraction on PGDP5K. Table \ref{primitive detection1} depicts the geometric primitive detection results using the evaluation manner 1, in which line instance covers all collinear line segments. We can see that our approach achieves a remarkable improvement over traditional methods such as Freeman \cite{Pratt2001} and GEOS \cite{Seo2015}, particularly on point and line. Appendix D lists the performance of all primitive classes adopting the evaluation manner 2. We find that most primitives could be well located and recognized except some minority classes, and joint classification in the GNN evidently improves the performance of text recognition compared with classification in the NDM.
       \begin{table}[t]
        \small
        \renewcommand\arraystretch{1.1}
        \centering
        \begin{tabular}{cl|ccc}
        \toprule
           \multicolumn{2}{c|}{} & Freeman & GEOS & PGDPNet \\
        \midrule
          \multirow{3}*{Point}& Precision & 67.46 & 76.51 & \textbf{99.65}  \\
                              & Recall & 80.41 & 93.44 & \textbf{99.71}  \\
                              & F1 & 73.37 & 84.13 & \textbf{99.68} \\
        \midrule
          \multirow{3}*{Line}& Precision & 50.78 & 66.99 & \textbf{99.30} \\
                      & Recall &  80.43 & 90.46 & \textbf{99.51} \\
                      & F1 & 62.25  & 76.98 & \textbf{99.40} \\
        \midrule
          \multirow{3}*{Circle}& Precision & 90.72 & 98.25 & \textbf{99.85} \\
                      & Recall & 97.75 & 99.24 & \textbf{99.96} \\
                      & F1 & 94.10 &  98.74 & \textbf{99.90} \\
        \bottomrule
        \end{tabular}
        \caption{Detection performance of geometric primitives with the evaluation manner 1.}
        \label{primitive detection1}
    \end{table}

\subsection{Primitive Relation Parsing}
To better demonstrate advantages of graph feature generation, we conducted ablation studies in primitive relation parsing. Table \ref{primitive relation construction} displays performances of different feature initialization methods, where baseline denotes the method of with only visual-location embedding, SE and PL refer to class semantic feature and parsing position feature formulated in Eq (\ref{eqVL}), respectively. The performance gap is mainly reflected in the relationships of text2geo, sym2geo and text2head. However, most relations among primitives belong to geo2geo, so the overall performance of relationships shows little difference. We also compare the methods using another evaluation indicator \textit{complete accuracy}, which refers to the proportion of complete correct sample. On the whole, fusion with parsing position and class semantic information makes model easier to learn representative features so as to promote relation reasoning, and gains 1.7\% complete accuracy improvement.

    \begin{table}[t]
        \scriptsize
        \renewcommand\arraystretch{1.2}
        \centering
        \begin{tabular}{cl|cccc}
        \toprule
           \multicolumn{2}{c|}{} & Baseline & w SE & w PL & w SE\&PL \\
        \midrule 
          \multirow{3}*{All}& Precision & 98.96 & 99.16 & 99.10 & \textbf{99.16}\\
                              & Recall & 96.97 & 97.01 & \textbf{97.11} & 97.07\\
                              & F1 & 97.96 & 98.07 & 98.08 & \textbf{98.11}\\
        \midrule
          \multirow{3}*{Geo2Geo}& Precision & 98.84 & \textbf{99.15} & 99.09 & 99.13\\
                              & Recall & 98.53 & 98.56 & \textbf{98.69} & 98.60 \\
                              & F1 & 98.68 & 98.85 & \textbf{98.89} & 98.86 \\
        \midrule 
          \multirow{3}*{Text2Geo}& Precision & 99.09 & 99.27 & \textbf{99.38} & 99.27 \\
                              & Recall & 96.16 & 96.61 & 96.36 & \textbf{96.84}\\
                              & F1 & 97.60 & 97.94 & 97.84 & \textbf{98.04}\\
        \midrule  
          \multirow{3}*{Sym2Geo}& Precision & 99.06 & 98.71 & 99.01 & \textbf{99.07} \\
                              & Recall & 94.13 & 94.89 & 95.02 & \textbf{95.27} \\
                              & F1 & 96.53 & 96.76 & 96.97 & \textbf{97.13} \\
        \midrule  
          \multirow{3}*{Text2Head}& Precision & 97.70 & 98.03 & 98.03 & \textbf{98.08} \\
                              & Recall & 91.95 & 92.26 & 92.57 & \textbf{95.05} \\
                              & F1 & 94.74 & 95.06 & 95.22 & \textbf{96.54} \\
        \midrule
          \multicolumn{2}{c|}{Complete Acc}& 81.50 & 82.50 & 82.60 & \textbf{83.20}\\
        \bottomrule
        \end{tabular}
        \caption{Ablation studies of primitive relation parsing. "A2B" denotes the relationship between class A and class B by default.}
        \label{primitive relation construction}
    \end{table}

    
    
\subsection{Geometry Formal Language Generation}
We also conducted experiments in the generation of geometry formal language for a more advanced evaluation. The outcomes of geometry formal language depend on the complete results of primitive extraction and relation reasoning to form comprehensible geometric propositions, and any error in preceding sub-tasks will influence the final generation results. Table \ref{geometry formal language} shows experimental results of all, geo2geo and non-geo2geo relationship. Our method without GNN sharply outperforms the InterGPS on geo2geo relationship, and the GNN module further improves non-geo2geo relationship reasoning in spite of a slight performance decline on geo2geo, because segmentation results are already accurate enough to determine the geo2geo relation by the distance rules. Finally, on two datasets, our PGDPNet respectively achieves 57.2\% and 57.4\% improvements of Totally Same of All compared with the InterGPS, and exceeds the one without GNN by 11.0\% and 6.5\%.

     \begin{table}[t]
        \renewcommand\arraystretch{1.2}
        \centering
        \scriptsize
        \begin{tabular}{cl|ccc}
        \toprule
          \multicolumn{2}{c|}{} & IMP-Geometry3K & PGDP5K \\
        \midrule
          \multirow{4}*{All} & Likely Same & 73.71 / 99.17 / \textbf{99.33} & 65.70 / 98.40 / \textbf{99.00}\\
                              & Almost Same & 50.08 / 95.51 / \textbf{98.50}  & 44.40 / 93.10 / \textbf{96.60}\\ 
                              & Perfect Recall & 45.26 / 81.03 / \textbf{92.18} & 40.00 / 79.70 / \textbf{86.20}\\
                              & Totally Same & 34.28 / 80.53 / \textbf{91.51} & 27.30 / 78.20 / \textbf{84.70}\\
        \midrule
          \multirow{4}*{Geo2Geo}& Likely Same & 69.88 / \textbf{99.67} / 99.50 & 63.90 / \textbf{99.10} / 99.00 \\
                              & Almost Same & 56.24 / \textbf{99.50} / 99.00 & 49.40 / \textbf{97.30} / 97.10 \\
                              & Perfect Recall & 74.71 / \textbf{99.33} / 99.17 & 78.70 / 96.90 / \textbf{97.40}\\
                              & Totally Same & 47.59 / \textbf{98.84} / 98.33 & 40.80 / 93.60 / \textbf{94.50}\\
        \midrule
          \multirow{4}*{\makecell[c]{Non-geo\\2Geo}}& Likely Same & 77.04 / 96.01 / \textbf{99.00} & 67.30 / 95.80 / \textbf{98.00}\\
                              & Almost Same & 59.07 / 89.35 / \textbf{96.01} & 49.80 / 88.20 / \textbf{94.90}\\
                              & Perfect Recall & 50.92 / 81.20 / \textbf{92.85} & 45.70 / 81.30 / \textbf{87.00}\\
                              & Totally Same & 48.59 / 80.87 / \textbf{92.85} & 40.50 / 80.60 / \textbf{86.40}\\
        \bottomrule
        \end{tabular}
        \caption{Evaluation results of specification generation in geometry formal language. "\&/\&/\&" denotes performances of three methods compared: InterGPS, PGDPNet without GNN and PGDPNet.}
        \label{geometry formal language}
    \end{table}

\subsection{InterGPS System Problem Solving}
    To show the potential of our approach in geometry problem solving, we evaluate the performance using an existing problem solver, the one of InterGPS system, by replacing its geometry diagram parser with ours while remaining other modules unchanged. Table \ref{problem solving} reports the Inter-GPS performance feeding with different sources of propositions. When using the text parser of InterGPS with propositions generated from our PGDPNet, Inter-GPS achieves accuracy of 74.1\%, nearly 16.6\% higher than the diagram parser of InterGPS. The GM improves performance by 4.8\% compared to the one without GNN. Slight gaps among generated diagram propositions, generated text propositions and annotations show that the symbolic geometry solver of InterGPS still has much room to improve. 
    \begin{table}[t]
        \small
        \renewcommand\arraystretch{1.1}
        \centering
        \begin{tabular}{l|cc}
        \toprule
          & Text InterGPS & Text GT \\
        \midrule
          Diagram w/o & 25.4$\pm$0.0 & 25.4$\pm$0.0\\
          Diagram InterGPS & 57.5$\pm$0.2 & 58.0$\pm$1.7\\
          Diagram PGDPNet w/o GNN & 69.3$\pm$0.2 & 70.0$\pm$0.4 \\
          Diagram PGDPNet & 74.1$\pm$0.2 & 74.3$\pm$0.3 \\
          Diagram GT & \textbf{75.9$\pm$0.2} & \textbf{76.0$\pm$0.4} \\
        \bottomrule
        \end{tabular}
        \caption{Problem solving accuracy of InterGPS system on IMP-Geometry3K dataset.}
        \label{problem solving}
    \end{table}

\subsection{Limitations}
\begin{figure}[t]
        \begin{center}
        \includegraphics[width=1.0 \columnwidth,trim=10 5 5 5,clip]{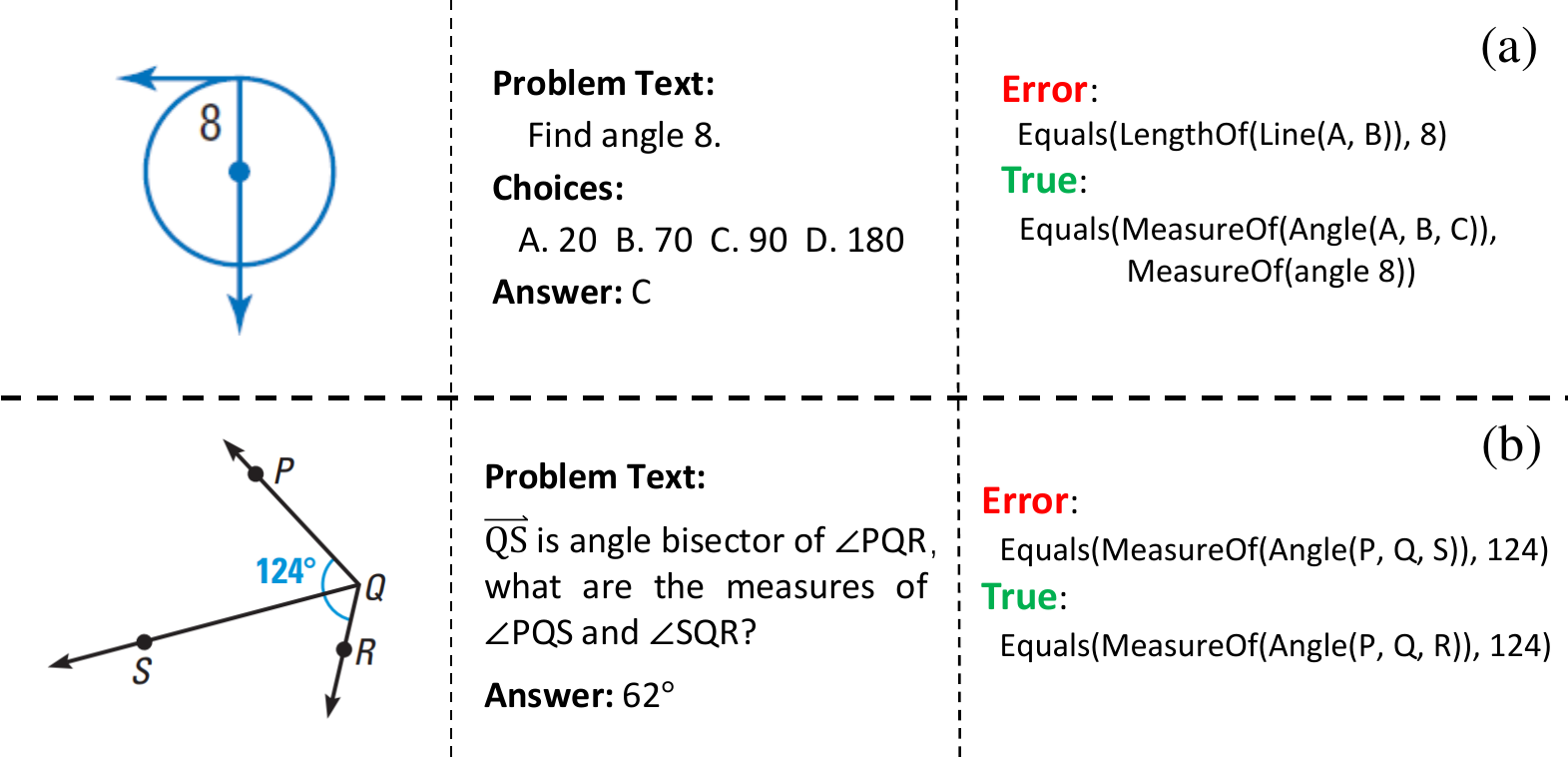} 
        \end{center}
        \caption{Failure examples of our method.}
        \label{limitation}
    \end{figure}

    We show some failure cases of our method in Figure \ref{limitation}. In Figure \ref{limitation}(a), the text “$8$" is mistaken as the radius of circle, while the problem text shows that it is an angle label. In Figure \ref{limitation}(b), The text “$124^{\circ}$" is incorrectly denoted as the degree of $\angle PQS$ but is actually the degree of $\angle PQR$. This reveals that geometry diagram parsing should not rely on images alone but also make full use of textual semantics, and it even involves geometry logical reasoning. Future works will consider incorporating the text description to aid diagram parsing to further improve parsing performance.
    
\section{Conclusion}
We propose the first end-to-end deep learning model PGDPNet for PGDP, which gives explicit primitive instance extraction, classification and between-primitive relationship reasoning. We also construct a new large-scale geometry diagram dataset PGDP5K with primitive level annotations. Experimental results demonstrate the superiority of proposed parsing method. This work promotes the benchmark of plane geometry diagram parsing, and provides a powerful tool to aid geometry problem solving and Q\&A.

\section*{Acknowledgments}
We thank Yunfei Guo, Jinwen Wu, and Xiaolong Yun for helpful discussions. This work has been supported by the National Key Research and Development Program under Grant No. 2020AAA0109702, the National Natural Science Foundation of China (NSFC) grants 61733007, 61721004.
 
\bibliographystyle{named}
\bibliography{ijcai22}

\begin{thebibliography}{}

\bibitem[\protect\citeauthoryear{Amini \bgroup \em et al.\egroup
  }{2019}]{Amini2019}
Aida Amini, Saadia Gabriel, Shanchuan Lin, Rik Koncel-Kedziorski, Yejin Choi,
  and Hannaneh Hajishirzi.
\newblock {MathQA: Towards interpretable math word problem solving with
  operation-based formalisms}.
\newblock In {\em NAACL HLT}, 2019.

\bibitem[\protect\citeauthoryear{Chen \bgroup \em et al.\egroup
  }{2021}]{Chen2021}
Jiaqi Chen, Jianheng Tang, Jinghui Qin, Xiaodan Liang, Lingbo Liu, Eric Xing,
  and Liang Lin.
\newblock {GeoQA: A geometric question answering benchmark towards multimodal
  numerical reasoning}.
\newblock In {\em Findings of ACL}, 2021.

\bibitem[\protect\citeauthoryear{Chou \bgroup \em et al.\egroup
  }{1996}]{Chou1996}
Shang~Ching Chou, Xiao~Shan Gao, and Jing~Zhong Zhang.
\newblock Automated generation of readable proofs with geometric invariants:
  {II}. theorem proving with full-angles.
\newblock {\em Journal of Automated Reasoning}, 17:349--370, 1996.

\bibitem[\protect\citeauthoryear{{De Brabandere} \bgroup \em et al.\egroup
  }{2017}]{DeBrabandere2017}
Bert {De Brabandere}, Davy Neven, and Luc {Van Gool}.
\newblock {Semantic instance segmentation with a discriminative loss function}.
\newblock In {\em CoRR}, volume abs/1708.0, 2017.

\bibitem[\protect\citeauthoryear{Gan \bgroup \em et al.\egroup
  }{2018}]{Gan2018}
Wenbin Gan, Xinguo Yu, Chao Sun, Bin He, and Mingshu Wang.
\newblock Understanding plane geometry problems by integrating relations
  extracted from text and diagram.
\newblock {\em PSIVT 2017: Image and Video Technology, LNCS}, 10749:366--381,
  2018.

\bibitem[\protect\citeauthoryear{Guo \bgroup \em et al.\egroup
  }{2021}]{Guo2021}
Yunfei Guo, Wei Feng, Fei Yin, Tao Xue, Shuqi Mei, and Cheng-Lin Liu.
\newblock {Learning to understand traffic signs}.
\newblock In {\em ACM MM}, 2021.

\bibitem[\protect\citeauthoryear{He \bgroup \em et al.\egroup }{2018}]{He2018}
Kaiming He, Georgia Gkioxari, Piotr Dollár, and Ross Girshick.
\newblock Mask r-cnn.
\newblock {\em IEEE Transactions on Pattern Analysis and Machine Intelligence},
  42:386--397, 2018.

\bibitem[\protect\citeauthoryear{Huang \bgroup \em et al.\egroup
  }{2018}]{Huang2018}
Kun Huang, Yifan Wang, Zihan Zhou, Tianjiao Ding, Shenghua Gao, and Yi~Ma.
\newblock Learning to parse wireframes in images of man-made environments.
\newblock In {\em CVPR}, 2018.

\bibitem[\protect\citeauthoryear{Lin \bgroup \em et al.\egroup
  }{2017}]{Lin2017}
Tsung~Yi Lin, Piotr Doll{\'{a}}r, Ross Girshick, Kaiming He, Bharath Hariharan,
  and Serge Belongie.
\newblock {Feature pyramid networks for object detection}.
\newblock In {\em CVPR}, 2017.

\bibitem[\protect\citeauthoryear{Liu \bgroup \em et al.\egroup
  }{2021}]{Liu2021}
Hengyue Liu, Ning Yan, Masood Mortazavi, and Bir Bhanu.
\newblock {Fully convolutional scene graph generation}.
\newblock In {\em CVPR}, 2021.

\bibitem[\protect\citeauthoryear{Lu \bgroup \em et al.\egroup }{2021}]{Lu2021}
Pan Lu, Ran Gong, Shibiao Jiang, Liang Qiu, Siyuan Huang, Xiaodan Liang, and
  Song-Chun Zhu.
\newblock {Inter-GPS: Interpretable geometry problem solving with formal
  language and symbolic reasoning}.
\newblock In {\em ACL-IJCNLP}, 2021.

\bibitem[\protect\citeauthoryear{Neven \bgroup \em et al.\egroup
  }{2018}]{Neven2018}
Davy Neven, Bert~De Brabandere, Stamatios Georgoulis, Marc Proesmans, and
  Luc~Van Gool.
\newblock {Towards end-to-end lane detection: An instance segmentation
  approach}.
\newblock {\em Proceedings of IEEE Intelligent Vehicles Symposium}, pages
  286--291, 2018.

\bibitem[\protect\citeauthoryear{Neven \bgroup \em et al.\egroup
  }{2019}]{Neven2019}
Davy Neven, Bert~De Brabandere, Marc Proesmans, and Luc {Van Gool}.
\newblock {Instance segmentation by jointly optimizing spatial embeddings and
  clustering bandwidth}.
\newblock In {\em CVPR}, 2019.

\bibitem[\protect\citeauthoryear{Pratt}{2007}]{Pratt2001}
William~K Pratt.
\newblock {\em {Digital Image Processing, Fourth Edition}}.
\newblock Wiley, 2007.

\bibitem[\protect\citeauthoryear{Sachan \bgroup \em et al.\egroup
  }{2017}]{Sachan2017}
Mrinmaya Sachan, Avinava Dubey, and Eric~P. Xing.
\newblock {From textbooks to knowledge: A case study in harvesting axiomatic
  knowledge from textbooks to solve geometry problems}.
\newblock In {\em EMNLP}, 2017.

\bibitem[\protect\citeauthoryear{Sachan \bgroup \em et al.\egroup
  }{2020}]{Sachan2020}
Mrinmaya Sachan, Avinava Dubey, Eduard~H. Hovy, Tom~M. Mitchell, Dan Roth, and
  Eric~P. Xing.
\newblock {Discourse in multimedia: A case study in extracting geometry
  knowledge from textbooks}.
\newblock {\em Computational Linguistics}, 45:627--665, 2020.

\bibitem[\protect\citeauthoryear{Sandler \bgroup \em et al.\egroup
  }{2018}]{Sandler2018}
Mark Sandler, Andrew Howard, Menglong Zhu, Andrey Zhmoginov, and Liang~Chieh
  Chen.
\newblock {MobileNetV2: Inverted residuals and linear bottlenecks}.
\newblock In {\em CVPR}, 2018.

\bibitem[\protect\citeauthoryear{Seo \bgroup \em et al.\egroup
  }{2014}]{Seo2014}
Min~Joon Seo, Hannaneh Hajishirzi, Ali Farhadi, and Oren Etzioni.
\newblock {Diagram understanding in geometry questions}.
\newblock In {\em AAAI}, 2014.

\bibitem[\protect\citeauthoryear{Seo \bgroup \em et al.\egroup
  }{2015}]{Seo2015}
Minjoon Seo, Hannaneh Hajishirzi, Ali Farhadi, Oren Etzioni, and Clint Malcolm.
\newblock {Solving geometry problems: Combining text and diagram
  interpretation}.
\newblock In {\em EMNLP}, 2015.

\bibitem[\protect\citeauthoryear{Tian \bgroup \em et al.\egroup
  }{2020}]{Tian2020}
Zhi Tian, Chunhua Shen, Hao Chen, and Tong He.
\newblock {FCOS: A simple and strong anchor-free object detector}.
\newblock {\em IEEE Transactions on Pattern Analysis and Machine Intelligence},
  8828:1--1, 2020.

\bibitem[\protect\citeauthoryear{Veli{\v{c}}kovi{\'{c}} \bgroup \em et
  al.\egroup }{2018}]{Romero2018}
Petar Veli{\v{c}}kovi{\'{c}}, Arantxa Casanova, Pietro Li{\`{o}}, Guillem
  Cucurull, Adriana Romero, and Yoshua Bengio.
\newblock {Graph attention networks}.
\newblock In {\em ICLR}, 2018.

\bibitem[\protect\citeauthoryear{Xu \bgroup \em et al.\egroup }{2017}]{Xu2017}
Danfei Xu, Yuke Zhu, Christopher~B. Choy, and Li~Fei-Fei.
\newblock Scene graph generation by iterative message passing.
\newblock In {\em CVPR}, 2017.

\bibitem[\protect\citeauthoryear{Ye \bgroup \em et al.\egroup }{2020}]{Ye2020}
Jun-Yu Ye, Yan-Ming Zhang, Qing Yang, and Cheng-Lin Liu.
\newblock Contextual stroke classification in online handwritten documents with
  edge graph attention networks.
\newblock {\em SN Computer Science}, 1(3), 2020.

\bibitem[\protect\citeauthoryear{Ying \bgroup \em et al.\egroup
  }{2021}]{Ying2021}
Hui Ying, Zhaojin Huang, Shu Liu, Tianjia Shao, and Kun Zhou.
\newblock Embedmask: Embedding coupling for one-stage instance segmentation.
\newblock In {\em IJCAI}, 2021.

\bibitem[\protect\citeauthoryear{Zhou \bgroup \em et al.\egroup
  }{2019}]{Zhou2019}
Yichao Zhou, Haozhi Qi, and Yi~Ma.
\newblock End-to-end wireframe parsing.
\newblock In {\em ICCV}, 2019.

\end{thebibliography}

\newpage
\appendix
\section*{Appendix} 
\vspace{0.2cm}
\section{Primitive Examples}
\begin{figure}[htbp]
        \begin{center}
        \includegraphics[width=1.0\columnwidth,trim=10 30 10 0,clip]{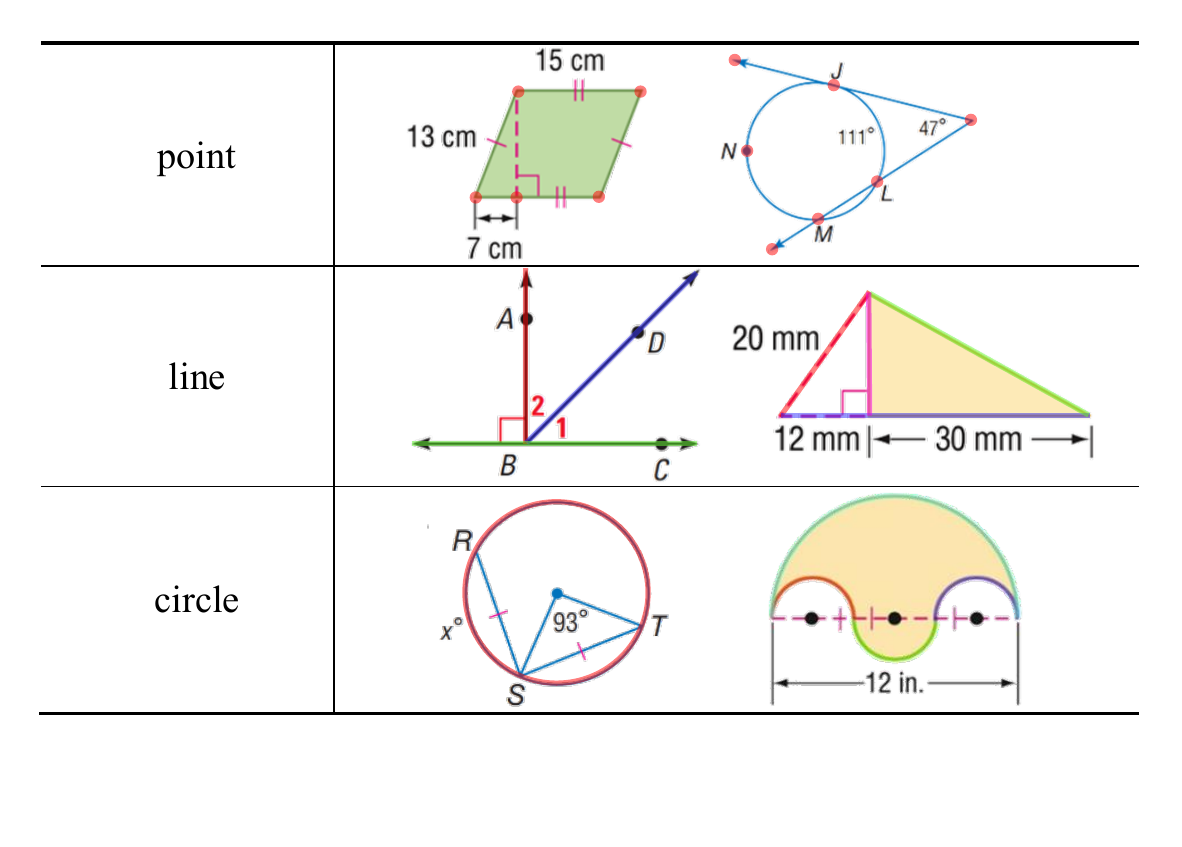} 
        \end{center}
        \vspace{-0.3cm}
        \caption{Examples of geometric primitive.}
        \label{geometric primitive}
    \end{figure}

\begin{figure}[H]
        \begin{center}
        \includegraphics[width=1.0\columnwidth,trim=10 25 10 0,clip]{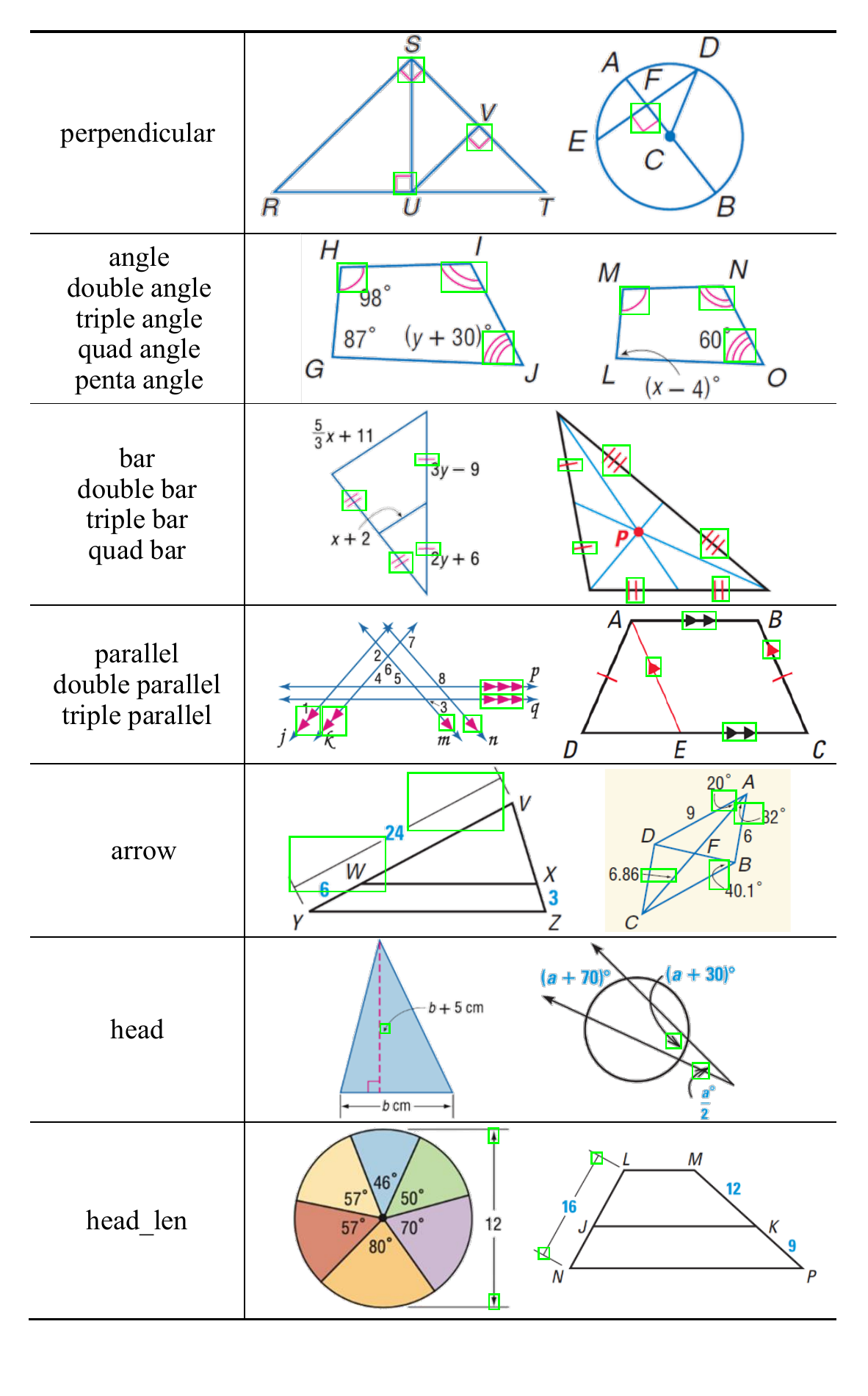} 
        \end{center}
        \vspace{-0.3cm}
        \caption{Examples of symbol.}
        \label{symbol}
    \end{figure}

\begin{figure}[H]
        \begin{center}
        \includegraphics[width=1.0\columnwidth,trim=10 30 10 0,clip]{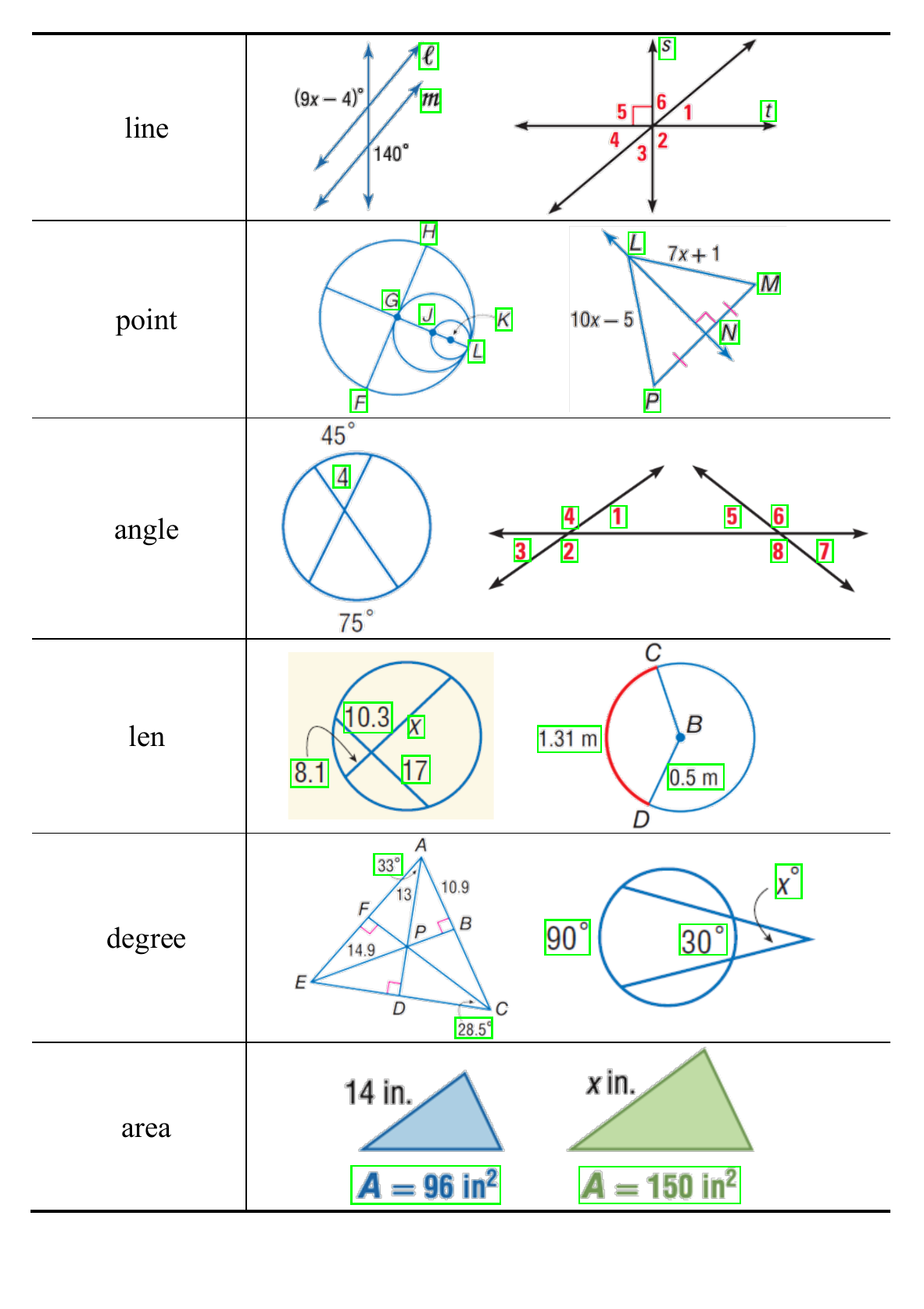} 
        \end{center}
        \vspace{-0.3cm}
        \caption{Examples of text.}
        \label{text}
    \end{figure}

    Several typical examples of \textit{geometric primitive}, \textit{symbol} and \textit{text}  are displayed in Figure \ref{geometric primitive}, Figure \ref{symbol} and Figure \ref{text}, respectively:
    \begin{itemize}
    	\item[$\bullet$] The geometric primitive includes \textit{point}, \textit{line} and \textit{circle} three classes. The point covers intersection point, tangent point, endpoint and independent point. The line consists of solid line, dash line and mixture of solid and dash. It is worth noting that we only label the longest line segment of all collinear lines. The circle includes complete circle and arc.
	    \item[$\bullet$] Symbol has 6 super-classes and 16 sub-classes: \textit{perpendicular}, \textit{angle}, \textit{bar}, \textit{parallel}, \textit{arrow} and \textit{head}, where super-classes of angle, bar and parallel have multiple sub-classes, and head is subdivided into two sub-classes to distinguish the indication relations of different types of arrows.
    	\item[$\bullet$] We divide text into 6 classes, including \textit{line}, \textit{point}, \textit{angle}, \textit{length}, \textit{degree} and \textit{area}. In many cases, there is no visual distinction among different text classes, $e.g.$ angle, length and degree. So we need to combine spatial and structure information to carry out the fine-grained classification of text. 
	\end{itemize}

    To sum up, the diagrams in PGDP5K have more complicated layouts and even primitives of the same class have great difference in style, which make our dataset more challenging. 
    
    \section{Dataset PGDP5K Distribution} 
\begin{figure*}[htbp]
        \begin{center}
        \includegraphics[width=2.0\columnwidth]{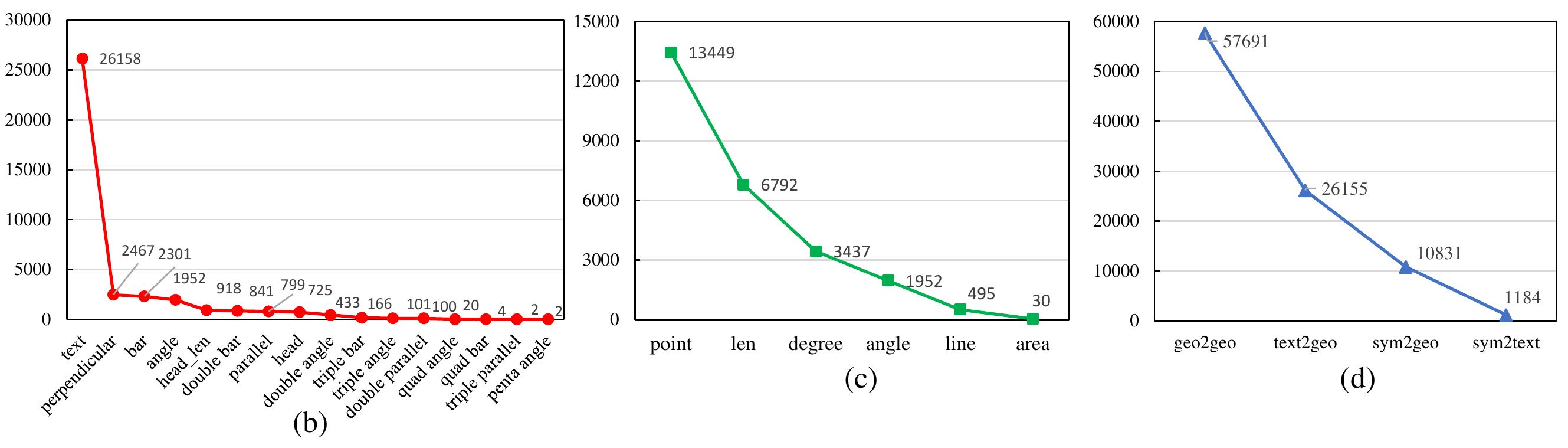} 
        \end{center}
        \vspace{-0.4cm}
        \caption{Distributions of PGDP5K Dataset. (a)(b)(c)(d) respectively denote the class distribution of shape,  symbol, text and relation.}
        \label{dataset distribution}
    \end{figure*}

    Figure \ref{dataset distribution} displays class distributions of geometry shape, symbol, text and relation. They all obey the long-tailed distribution evidently. Note that text is seen as a special symbol recorded in the symbol distribution. In experiments, the minority class with few samples performs poorly in tasks of detection, classification and relation reasoning. Mitigating the effects of the long tail is a considerable direction for the future research of PGDP.
    
    \section{Geometric Proposition Templates}
    We realize geometric propositions about basic primitive relations listed in Table \ref{geometric proposition templates}: geometry shape, geometric primitive with geometric primitive, text with geometric primitive, and symbol with geometric primitive:
    
    \begin{itemize}
    	\item[$\bullet$] \vspace{-0.1cm} Geometry shapes are basic elements of high-level propositions. We give proposition templates of 5 types of fundamental geometry shapes: \textit{point}, \textit{line}, \textit{circle}, \textit{angle} and \textit{arc}, where line, angle and arc have several equivalent expressions. 
	    \item[$\bullet$] \vspace{-0.1cm} We define 3 types of proposition templates about relations among geometric primitives: point lies on line, point lies on circle, and point is center of circle. These three primary relations could produce more other high-level relations among geometric primitives.  
    	\item[$\bullet$] \vspace{-0.1cm} According to text class, we divide relations of text with geometric primitive into 6 types of propositions, where proposition templates of degree and length are not unique. 
    	\item[$\bullet$] \vspace{-0.1cm} Same as text with geometric primitive, propositions of symbol with geometric primitive are divided into 4 groups due to symbol class, and there are 2 proposition templates of bar.
	\end{itemize}
	
	The design of proposition templates does not only relate to primitive relation parsing but also applies to logical reasoning of problem solving. Consequently, geometric proposition templates are the crystallization of geometry knowledge.
	
	\section{Primitive Detection Performance} 
    Table \ref{primitive detection2 detailed} exhibits performance results of each primitive class with evaluation manner 2 on PGDP5K. Since that evaluation manner 2 only applies to the segmentation route, in this experiment, we compare with methods of PGDPNet with or without the GNN module, but exclude the InterGPS method. The method of PGDPNet without GNN identifies the fine-grained text classes along with symbol classes in the NDM, while the PGDPNet first detects a coarse-grained text class and then implements fine-grained text classification in the GM, combined with visual, semantic and structural information. End-to-end learning with all modules not only promotes the primitive relation parsing but also slightly boosts performance of primitive extraction according to experimental results. Nevertheless, some minority classes with fewer samples, $e.g.$ "quad bar", "triple parallel" and "penta angle", perform poorly. 
    
    \begin{figure}[htbp]
        \begin{center}
        \includegraphics[width=0.8\columnwidth]{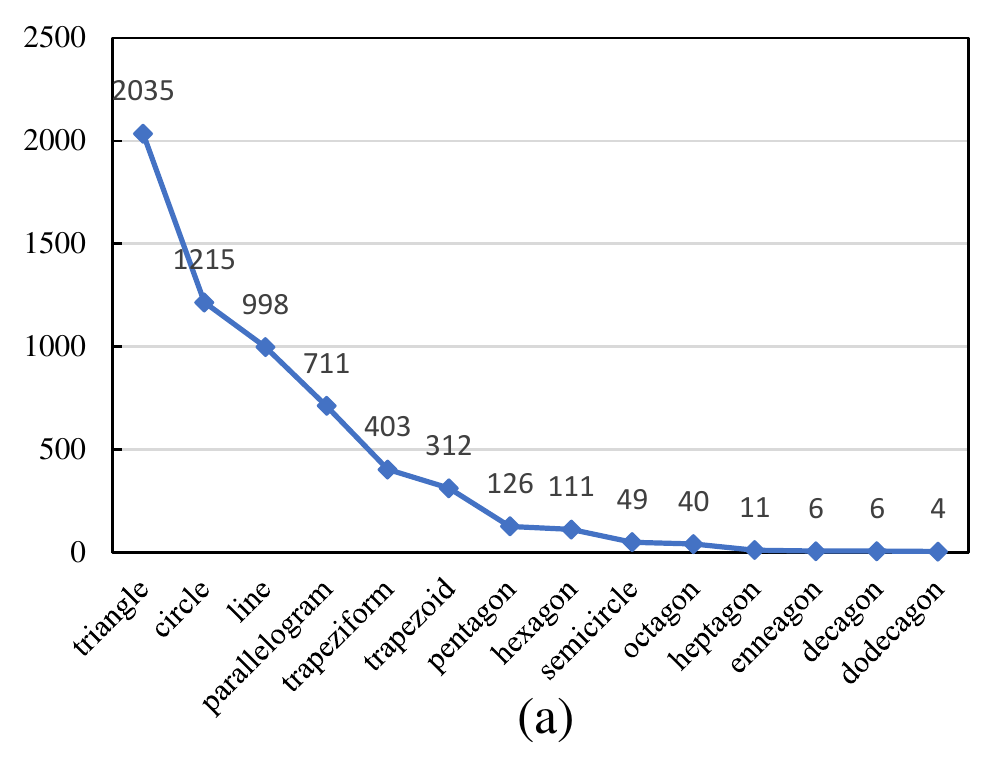} 
        \end{center}
    \end{figure}
    
    \newpage

    \newpage
    
    \begin{table*}[htbp]
        \renewcommand\arraystretch{0.8}
        \centering
        \footnotesize
        \begin{tabular}{c|c|l}
        \toprule
         \multirow{2.0}*{Relation Class}  & \multirow{2.0}*{Primitive Class} & \multirow{2.0}*{Proposition Templates} \\
          &  &  \\
        \midrule
        \multirow{7}*{Geo Shape}& Point & Point(\$)\\
            \cmidrule{2-3}
        	& Line & Line(\$, \$), Line(\$)\\
        	\cmidrule{2-3}
        	& Circle & Circle(\$, radius\_\$)\\
        	\cmidrule{2-3}
        	& Angle & Angle(\$, \$, \$), Angle(\$)\\
        	\cmidrule{2-3}
        	& Arc & Arc(\$, \$), Arc(\$, \$, \$)\\
        \midrule
        \multirow{3}*{Geo2Geo}& \multirow{3}*{Point} & PointLiesOnLine(\$, Line(\$, \$)) \\
              &       & PointLiesOnCircle(\$, Circle(\$, radius\_\$))\\
              &       & Circle(\$, radius\_\$)\\
        \midrule
        \multirow{10}*{Text(\&)2Geo}& Text\_point & Point(\&) \\
            \cmidrule{2-3}
            & Text\_line & Line(\&) \\
            \cmidrule{2-3}
            & Text\_angle & Equals(MeasureOf(Angle(\$, \$, \$)), MeasureOf(angle \&))\\
            \cmidrule{2-3}
            & \multirow{2}*{Text\_degree} & Equals(MeasureOf(Angle(\$, \$, \$)), \&)\\
            &             & Equals(MeasureOf(Arc(\$, \$)), \&)\\
            \cmidrule{2-3}
            & \multirow{2}*{Text\_len} & Equals(LengthOf(Line(\$, \$)), \&)\\
            &           & Equals(LengthOf(Arc(\$, \$)), \&)\\
            \cmidrule{2-3}
            & Text\_area & - \\
        \midrule
        \multirow{6}*{Sym2Geo}& Sym\_perpendicular & Perpendicular(Line(\$, \$), Line(\$, \$)) \\
            \cmidrule{2-3}
            & Sym\_angle & Equals(MeasureOf(Angle(\$, \$, \$)), MeasureOf(Angle(\$, \$, \$))) \\
            \cmidrule{2-3}
            &  \multirow{2}*{Sym\_bar} & Equals(LengthOf(Line(\$, \$)), LengthOf(Line(\$, \$)))\\
            &                          & Equals(LengthOf(Arc(\$, \$)), LengthOf(Arc(\$, \$)))\\
            \cmidrule{2-3}
            & Sym\_parallel & Parallel(Line(\$, \$), Line(\$, \$)) \\
        \bottomrule
        \end{tabular}
        \vspace{-0.1cm}
        \caption{Geometric proposition templates of primitive relation.}
        \vspace{-0.04cm}
        \label{geometric proposition templates}
    \end{table*}

    \begin{table*}[htbp]
        \renewcommand\arraystretch{0.95}
        \centering
        \footnotesize
        \begin{tabular}{l|p{1.2cm}<{\centering}p{1.2cm}<{\centering}p{1.2cm}<{\centering}p{1.2cm}<{\centering}p{1.2cm}<{\centering}p{1.2cm}<{\centering}}
        \toprule
         \multirow{2.0}*{}  & \multicolumn{3}{c}{PGDP-Net w/o GNN} &  \multicolumn{3}{c}{PGDP-Net} \\
         \cmidrule(l){2-4} \cmidrule(l){5-7}
          & Precision & Recall & F1 & Precision & Recall & F1\\
        \midrule
          Geo-All & 99.57 & 99.49 & 99.53  & 99.55 & 99.36 & 99.46 \\
          point   & 99.44 & 99.60 & 99.52 & 99.46 &  99.48 & 99.47 \\
          line  & 99.77 & 99.40 & 99.58 & 99.70 & 99.31 & 99.50 \\
          circle  & 99.31 & 98.63 & 98.97 & 99.31 & 97.95 & 98.63 \\
        \midrule
          Sym-All  & 99.51 & 99.70 & 99.61 & 99.57 & 99.73 & 99.65 \\ 
          text & - & - & - & 99.70 & 99.96 & 99.83 \\
          perpendicular & 99.59 & 99.79 & 99.69 & 99.79 & 99.59 & 99.69 \\
          head  & 97.93 & 99.30 & 98.61 & 99.30 & 98.60 & 98.95 \\ 
          head\_len  & 100.00 & 97.80 & 98.89 & 100.00 & 98.90 & 99.45 \\ 
          angle  & 99.26 & 98.17 & 98.71 & 98.90 & 98.53 & 98.72 \\
          double angle  & 96.94 & 98.96 & 97.94 & 95.96 & 98.96 & 97.44\\
          triple angle  & 100.00 & 94.12 & 96.97 & 100.00 & 97.06 & 98.51 \\
          quad angle  & 80.00 & 100.00	& 88.89 & 100.00 & 75.00 & 85.71 \\
          penta angle  & 0.00 & 0.00 & 0.00 & 0.00 & 0.00 & 0.00\\
          bar  &  98.68 & 99.56 & 99.12 & 98.90	& 99.78 & 99.34\\
          double bar  & 100.00 & 100.00 & 100.00 & 100.00 & 99.43 & 99.72 \\
          triple bar  & 100.00 & 100.00 & 100.00 & 100.00 & 100.00 & 100.00\\
          quad bar  & 0.00 & 0.00 & 0.00 & 0.00 & 0.00 & 0.00 \\
          parallel  & 100.00 & 100.00 & 100.00 & 95.86 & 100.00 & 97.89\\
          double parallel  & 100.00 & 94.44 & 97.14 & 100.00 & 100.00 & 100.00\\
          triple parallel  & 0.00 & 0.00 & 0.00 & 0.00 & 0.00 & 0.00\\
          
        \midrule
          Text-All  & - & - & - & 99.55 & 99.81 & 99.68 \\
          text\_point & 99.93 & 99.86 & 99.89 & 99.86 & 99.82 & 99.84 \\
          text\_line  & 99.09 & 99.09 & 99.09 & 98.21 & 100.00 & 99.10\\ 
          text\_len  & 98.94 & 99.92 & 99.43 & 98.94 & 99.77 & 99.35\\ 
          text\_angle  & 99.71 & 100.00 & 99.86 & 99.43 & 99.71 & 99.57\\ 
          text\_degree  & 99.58 & 99.86 & 99.72 & 99.72 & 99.86 & 99.79\\ 
          text\_area  & 100.00 & 100.00 & 100.00 & 100.00 & 100.00 & 100.00\\
        \bottomrule
        \end{tabular}
        \vspace{-0.1cm}
        \caption{Detailed performance of primitive detection with the evaluation manner 2.}
        \vspace{-0.1cm}
        \label{primitive detection2 detailed}
    \end{table*}
\end{document}


\maketitle

\appendix

\section{Primitive Examples}
    Several typical examples of \textit{geometric primitive}, \textit{symbol} and \textit{text}  are respectively displayed in Figure \ref{geometric primitive}, Figure \ref{symbol} and Figure \ref{text} :
    \begin{itemize}[leftmargin=0.4cm]
    	\item[$\bullet$] The geometric primitive includes \textit{point}, \textit{line} and \textit{circle} three classes. The point covers intersection point, tangent point, endpoint and independent point, The line consists of solid line, dash line and mixture of solid and dash. It's worth noting that we only label the longest line segment of all collinear lines. The circle includes complete circle and arc.
	    \item[$\bullet$] Symbol has 6 super-classes and 16 sub-classes: \textit{perpendicular}, \textit{angle}, \textit{bar}, \textit{parallel}, \textit{arrow} and \textit{head}, where classes of angle, bar and parallel have multiple forms, and head is subdivided into two classes to distinguish different arrow indication relations.
    	\item[$\bullet$] We divide text into 6 classes, including \textit{line}, \textit{point}, \textit{angle}, \textit{len}, \textit{degree} and \textit{area}. In many cases, there is no visual distinction between different text classes, $e.g.$, angle, len and degree. As to fine-grained classification of text, we need to combine with spatial and structure information.
	\end{itemize}

    To sum up, the diagram in PGDP5K has more complicated layouts and even primitives of same class have great difference in style, which make our dataset more challenging.

    \begin{figure} [htbp]
        \begin{center}
        \includegraphics[width=1.0\columnwidth,trim=10 30 10 0,clip]{./pictures/primitive_class_geo.pdf} 
        \end{center}
        \vspace{-0.3cm}
        \caption{Examples of geometric primitive}
        \label{geometric primitive}
    \end{figure}
    
    \begin{figure}[htbp]
        \begin{center}
        \includegraphics[width=1.0\columnwidth,trim=10 25 10 0,clip]{./pictures/primitive_class_sym.pdf} 
        \end{center}
        \vspace{-0.2cm}
        \caption{Examples of symbol}
        \label{symbol}
    \end{figure}
    
    \begin{figure*}[htbp]
        \begin{center}
        \includegraphics[width=2.0\columnwidth]{./pictures/dataset distribution.pdf} 
        \end{center}
        \vspace{-0.4cm}
        \caption{Distributions of PGDP5K Dataset. (a)(b)(c) respectively denote the class distribution of geometry shape,  symbol and text.}
        \vspace{-0.4cm}  
        \label{dataset distribution}
    \end{figure*}
    
    \begin{figure}[H]
        \begin{center}
        \includegraphics[width=1.0\columnwidth,trim=10 30 10 0,clip]{./pictures/primitive_class_text.pdf} 
        \end{center}
        \vspace{-0.3cm}
        \caption{Examples of text}
        \label{text}
    \end{figure}
    
\vspace{-0.5cm}
\section{Dataset PGDP5K Distribution} 
    Figure \ref{dataset distribution} displays class distributions of geometry shape, symbol and text. They all obey the long-tailed distribution evidently. Note that text is seen as a special symbol recorded in the symbol distribution. In experiments, the minority class with few samples performs poorly in tasks of detection, classification and relation reasoning. 
  
\section{Geometric Proposition Templates}
    We realize geometry propositions about basic primitive relation listed in Table \ref{geometric proposition templates}, propositions of geometry shape, geometric primitive with geometric primitive, text with geometric primitive and symbol with geometric primitive:
    
    \begin{itemize}[leftmargin=0.4cm]
    	\item[$\bullet$] \vspace{-0.1cm} Geometry shapes are basic elements of high-level propositions. We give proposition templates of five types of fundamental geometry shapes: \textit{point}, \textit{line}, \textit{circle}, \textit{angle} and \textit{arc}, where line, angle and arc have several equivalent expressions. 
	    \item[$\bullet$] \vspace{-0.1cm}We define three types of proposition templates about relations among geometric primitives: point lies on line, point lies on circle and point is center of circle. These three primary relations could produce more other high-level relations of geometric primitives.  
    	\item[$\bullet$] \vspace{-0.1cm} According to text class, we divide into six types of propositions, where proposition templates of degree and len are not unique. 
    	\item[$\bullet$] \vspace{-0.1cm} Same as text with geometric primitive, propositions of symbol with geometric primitive are divided into four groups due to symbol class, and there are two proposition templates of bar.
	\end{itemize}
	
	The design of proposition templates does not only relate to primitive relation construction but also applies to symbolic reasoning of problem solving. Consequently, geometric proposition templates are the crystallization of geometry knowledge.
    
    \begin{table*}[htbp]
        \renewcommand\arraystretch{0.8}
        \centering
        \footnotesize
        \begin{tabular}{c|c|l}
        \toprule
         \multirow{2.0}*{Relation Class}  & \multirow{2.0}*{Primitive Class} & \multirow{2.0}*{Proposition Templates} \\
           &  &  \\
        \midrule
        \multirow{7}*{Geo Shape}& Point & Point(\$)\\
            \cmidrule{2-3}
        	& Line & Line(\$, \$), Line(\$)\\
        	\cmidrule{2-3}
        	& Circle & Circle(\$, radius\_\$)\\
        	\cmidrule{2-3}
        	& Angle & Angle(\$, \$, \$), Angle(\$)\\
        	\cmidrule{2-3}
        	& Arc & Arc(\$, \$), Arc(\$, \$, \$)\\
        \midrule
        \multirow{3}*{Geo2Geo}& \multirow{3}*{Point} & PointLiesOnLine(\$, Line(\$, \$)) \\
               &       & PointLiesOnCircle(\$, Circle(\$, radius\_\$))\\
               &       & Circle(\$, radius\_\$)\\
        \midrule
        \multirow{10}*{Text(\&)2Geo}& Text\_point & Point(\&) \\
            \cmidrule{2-3}
            & Text\_line & Line(\&) \\
            \cmidrule{2-3}
            & Text\_angle & Equals(MeasureOf(Angle(\$, \$, \$)), MeasureOf(angle \&))\\
            \cmidrule{2-3}
            & \multirow{2}*{Text\_degree} & Equals(MeasureOf(Angle(\$, \$, \$)), \&)\\
            &             & Equals(MeasureOf(Arc(\$, \$)), \&)\\
            \cmidrule{2-3}
            & \multirow{2}*{Text\_len} & Equals(LengthOf(Line(\$, \$)), \&)\\
            &           & Equals(LengthOf(Arc(\$, \$)), \&)\\
            \cmidrule{2-3}
            & Text\_area & - \\
        \midrule
        \multirow{6}*{Sym2Geo}& Sym\_perpendicular & Perpendicular(Line(\$, \$), Line(\$, \$)) \\
            \cmidrule{2-3}
            & Sym\_angle & Equals(MeasureOf(Angle(\$, \$, \$)), MeasureOf(Angle(\$, \$, \$))) \\
            \cmidrule{2-3}
            &  \multirow{2}*{Sym\_bar} & Equals(LengthOf(Line(\$, \$)), LengthOf(Line(\$, \$)))\\
            &                          & Equals(LengthOf(Arc(\$, \$)), LengthOf(Arc(\$, \$)))\\
            \cmidrule{2-3}
            & Sym\_parallel & Parallel(Line(\$, \$), Line(\$, \$)) \\
        \bottomrule
        \end{tabular}
        \vspace{-0.15cm}
        \caption{Geometric proposition templates of primitive relation}
        \vspace{-0.04cm}
        \label{geometric proposition templates}
    \end{table*}
    
\section{Primitive Detection Performance} 
    Table \ref{primitive detection2 detailed} exhibits results of each primitive class with evaluation manner 2 on PGDP5K. Since that evaluation manner 2 only applies to the segmentation route, in this experiment, we compare with our PGDPNet with or without GNN module but exclude the InterGPS method. The method of PGDPNet w/o GNN identifies the fine-grained text classes along with symbol classes in the NDM, while the PGDPNet first detects one coarse-grained text class and then carry fine-grained text classification in the GM combined with visual, semantic and structural information. End-to-end learning with all modules not only promotes the primitive relation construction but also slightly boosts performance of primitive extraction of many classes according to experimental results. Some classes with small amount of samples, $e.g.$, "quad bar", "triple parallel" and "penta angle", perform poorly. 

    \begin{table*}[htbp]
        \centering
        \footnotesize
        \begin{tabular}{l|p{1.2cm}<{\centering}p{1.2cm}<{\centering}p{1.2cm}<{\centering}p{1.2cm}<{\centering}p{1.2cm}<{\centering}p{1.2cm}<{\centering}}
        \toprule
         \multirow{2.0}*{}  & \multicolumn{3}{c}{PGDP-Net w/o GNN} &  \multicolumn{3}{c}{PGDP-Net} \\
         \cmidrule(l){2-4} \cmidrule(l){5-7}
           & Precision & Recall & F1 & Precision & Recall & F1\\
        \midrule
          Geo-All & 99.57 & 99.49 & 99.53  & 99.55 & 99.36 & 99.46 \\
          point   & 99.44 & 99.60 & 99.52 & 99.46 &  99.48 & 99.47 \\
          line  & 99.77 & 99.40 & 99.58 & 99.70 & 99.31 & 99.50 \\
          circle  & 99.31 & 98.63 & 98.97 & 99.31 & 97.95 & 98.63 \\
        \midrule
          Sym-All  & 99.51 & 99.70 & 99.61 & 99.57 & 99.73 & 99.65 \\ 
          text & - & - & - & 99.70 & 99.96 & 99.83 \\
          perpendicular & 99.59 & 99.79 & 99.69 & 99.79 & 99.59 & 99.69 \\
          head  & 97.93 & 99.30 & 98.61 & 99.30 & 98.60 & 98.95 \\ 
          head\_len  & 100.00 & 97.80 & 98.89 & 100.00 & 98.90 & 99.45 \\ 
          angle  & 99.26 & 98.17 & 98.71 & 98.90 & 98.53 & 98.72 \\
          double angle  & 96.94 & 98.96 & 97.94 & 95.96 & 98.96 & 97.44\\
          triple angle  & 100.00 & 94.12 & 96.97 & 100.00 & 97.06 & 98.51 \\
          quad angle  & 80.00 & 100.00	& 88.89 & 100.00 & 75.00 & 85.71 \\
          penta angle  & 0.00 & 0.00 & 0.00 & 0.00 & 0.00 & 0.00\\
          bar  &  98.68 & 99.56 & 99.12 & 98.90	& 99.78 & 99.34\\
          double bar  & 100.00 & 100.00 & 100.00 & 100.00 & 99.43 & 99.72 \\
          triple bar  & 100.00 & 100.00 & 100.00 & 100.00 & 100.00 & 100.00\\
          quad bar  & 0.00 & 0.00 & 0.00 & 0.00 & 0.00 & 0.00 \\
          parallel  & 100.00 & 100.00 & 100.00 & 95.86 & 100.00 & 97.89\\
          double parallel  & 100.00 & 94.44 & 97.14 & 100.00 & 100.00 & 100.00\\
          triple parallel  & 0.00 & 0.00 & 0.00 & 0.00 & 0.00 & 0.00\\
          
        \midrule
          Text-All  & - & - & - & 99.55 & 99.81 & 99.68 \\
          text\_point & 99.93 & 99.86 & 99.89 & 99.86 & 99.82 & 99.84 \\
          text\_line  & 99.09 & 99.09 & 99.09 & 98.21 & 100.00 & 99.10\\ 
          text\_len  & 98.94 & 99.92 & 99.43 & 98.94 & 99.77 & 99.35\\ 
          text\_angle  & 99.71 & 100.00 & 99.86 & 99.43 & 99.71 & 99.57\\ 
          text\_degree  & 99.58 & 99.86 & 99.72 & 99.72 & 99.86 & 99.79\\ 
          text\_area  & 100.00 & 100.00 & 100.00 & 100.00 & 100.00 & 100.00\\
        \bottomrule
        \end{tabular}
        \vspace{-0.1cm}
        \caption{Detailed performance of primitive detection with evaluation manner 2}
        \vspace{-0.1cm}
        \label{primitive detection2 detailed}
    \end{table*}